%% file: main.tex
\documentclass[letterpaper, 10 pt, conference]{ieeeconf}

\IEEEoverridecommandlockouts

\usepackage{amsmath,amssymb}
\usepackage{graphicx}
\usepackage[table]{xcolor}
\usepackage{booktabs}
\usepackage{tabularx}
\usepackage{array}
\usepackage{dcolumn}
\usepackage{cite}
\usepackage{placeins}
\usepackage{balance}
\newcolumntype{Y}{>{\centering\arraybackslash}X}
\newcolumntype{L}{>{\raggedright\arraybackslash}X}
\newcolumntype{d}[1]{D{.}{.}{#1}}
\definecolor{HiTacHeader}{RGB}{242,244,247}
\definecolor{HiTacDeploy}{RGB}{231,240,249}
\definecolor{HiTacSummary}{RGB}{246,246,246}

\title{\LARGE \bf
HiTac-WAM: A Hierarchical Tactile World Action Model for
Contact-Rich Robot Manipulation
}

\author{Chao Xue$^{1,2}$, Chaofan Zhang$^{1,2}$, Wenxuan Ma$^{1,2}$,
Guocai Yao$^{3}$, Shaowei Cui$^{1,2,3,*}$, and Shuo Wang$^{1}$\\
{\small $^{1}$Institute of Automation, Chinese Academy of Sciences}\\
{\small $^{2}$ImprintX Robotics}\\
{\small $^{3}$Beijing Academy of Artificial Intelligence}%
\thanks{$^{*}$Corresponding author: shaowei.cui@ia.ac.cn}}

\begin{document}

\maketitle
\thispagestyle{empty}
\pagestyle{empty}

\bstctlcite{BSTcontrol}

\begin{abstract}
World action models jointly predict future visual observations and actions,
whereas existing tactile-aware variants typically represent future touch as an
image or latent stream without modeling the physical dependencies that
organize tactile states hierarchically. We present HiTac-WAM, a
hierarchical tactile world action model that forecasts a sequence of future
tactile states for each candidate action chunk before execution. The forecast
factorizes into contact state, a 3D deformation field, and slip risk,
organized as a directed hierarchy in which each downstream stage is
conditioned on stop-gradient signals from preceding stages. A directed
attention mask allows tactile queries to attend to the video--action context
of each candidate while preventing video and action queries from attending to
tactile tokens. For planning, HiTac-WAM ranks candidate action chunks using
tactile forecasts and task-progress estimates. For execution, the selected
tactile forecast is retained as a reference; persistent discrepancies between
predicted and observed tactile states trigger corrective replanning. HiTac-WAM
achieves a mean contact F1 of 0.921; under matched training budgets, the
directed hierarchy reduces 3D displacement L2 error by 17.6\% relative to the
deformation-only predictor and improves slip AUPRC by 60.4\% relative to the
slip-only predictor. Across chip grasping, blackboard erasing, and USB
insertion, selection guided by the hierarchical forecasts increases the
average real-robot success rate from 31.1\% to 61.1\%, while the full system
attains 72.2\%.
\end{abstract}

\input{sections/01_introduction}
\input{sections/02_related_work}
\input{sections/03_method}
\input{sections/04_experiments}
\input{sections/05_results}
\input{sections/07_conclusion}

\balance
\bibliographystyle{IEEEtran}
\bibliography{references}

\end{document}

%% file: sections/01_introduction.tex
\section{Introduction}
\label{sec:introduction}

World action models jointly generate aligned future video and
continuous action chunks \cite{ye2026worldaction}. Contact-rich manipulation,
however, exposes a fundamental limitation of visual rollouts: contact at the
end-effector--environment interface is often only indirectly observable,
particularly under occlusion. This limitation becomes critical when repeated
sampling, conditioned on the same visual history, robot state, and language
instruction, uses independent noise draws to produce candidate action chunks
with paired visual rollouts. The
rollouts may appear similarly plausible---showing, for example, an aligned
gripper, a moving eraser, or a centered connector---while the corresponding
actions may lead to different physical outcomes, including missed or
unilateral contact, insufficient pressure, slip, and lateral jamming. The
central challenge is therefore to forecast the tactile consequences of each
candidate action chunk before execution.

\begin{figure}[!t]
  \centering
  \makebox[\linewidth][c]{%
    \includegraphics[width=\linewidth,trim=126bp 205bp 85bp 380bp,clip]{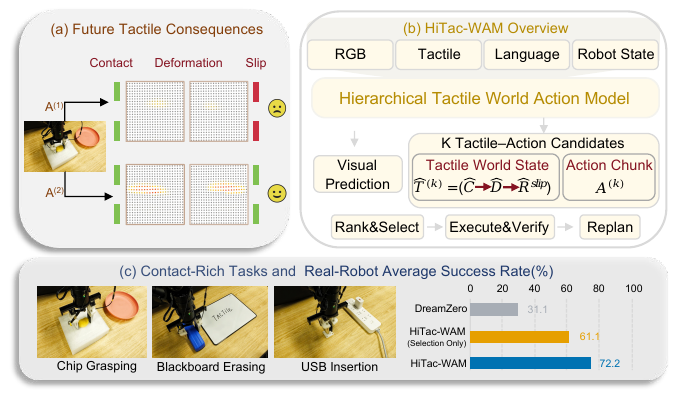}}
  \caption{\textbf{Overview of HiTac-WAM.}
  (a) Visually plausible candidate actions can produce distinct contact,
  deformation, and slip outcomes.
  (b) For each candidate action chunk, HiTac-WAM forecasts a hierarchical
  tactile trajectory ordered from contact state through deformation to slip
  risk, reflecting the physical dependencies among these states. It then
  ranks the resulting action--forecast pairs and retains the selected forecast
  as an execution-time reference for verification and replanning.
  (c) The three contact-rich tasks and observed average success rates in
  the main real-robot comparison.}
  \label{fig:motivation}
\end{figure}

Tactile policies condition actions on measured touch
\cite{chen2023vtt,huang2025threedvitac}, while slow-fast controllers provide
high-rate post-contact feedback \cite{xue2025reactive}; both respond only
after contact occurs. Predictive models instead forecast action-conditioned
changes in tactile observations, supporting model-predictive selection through
comparison with a goal tactile image \cite{tian2019manipulation} and enabling
prospective slip estimation \cite{mandil2022action}. Recent visuo-tactile
action models learn future tactile images or latents alongside policy features
\cite{heng2025vitacformer,lou2026dreamtac,zhang2026unitacvla}, demonstrating
that tactile forecasts can inform control. Existing predictive formulations,
however, typically either entangle contact state, interface deformation, and
slip risk within a single stream or treat them as independent targets, without
encoding their directed physical dependencies. Prior work has addressed
tactile forecasting, candidate evaluation, and online correction, but
generally in separate systems rather than through a unified mechanism that
ranks candidate action--forecast pairs using predictions structured by these
dependencies and retains the selected forecast as a temporally aligned
execution-time reference.

HiTac-WAM is built around three properties for tactile-forecast-based candidate
selection and execution-time verification: the forecast explicitly separates
the interface state into physically meaningful factors, each forecast remains
paired with its candidate action chunk during ranking, and the selected
forecast remains temporally aligned with subsequent observations. Accordingly,
the model (Fig.~\ref{fig:motivation}) represents the tactile consequences of
each candidate
action chunk as a structured trajectory, factorized into contact state,
contact-conditioned 3D deformation, and slip risk. The factors form a directed
hierarchy reflecting their physical dependencies: interface deformation is
meaningful only under contact, and slip risk depends on the resulting contact
and deformation states. Without explicit conditioning, a multi-output
predictor can capture these dependencies only implicitly; HiTac-WAM instead
passes upstream state signals to downstream stages through stop-gradient
conditioning. A directed attention mask further conditions each tactile
forecast on the corresponding candidate action chunk and its predicted visual
future within the same model. The model then ranks candidate action chunks
using the resulting tactile forecasts and task-progress estimates derived from
the predicted visual rollouts, and retains the selected forecast as an
execution-time reference; persistent forecast--observation deviations trigger
corrective actions.

Our main contributions are summarized as follows.
\begin{itemize}
\item We propose HiTac-WAM, a hierarchical tactile world action model that
factorizes the tactile forecast of each candidate action chunk into contact
state, contact-conditioned 3D deformation, and slip risk. Under matched
training budgets, the hierarchy reduces 3D displacement L2 error by $17.6\%$
relative to the deformation-only predictor and improves slip AUPRC by $60.4\%$
relative to the slip-only predictor.
\item We introduce forecast-guided action selection, which ranks candidate
action chunks by their tactile forecasts and task-progress estimates. On three
contact-rich tasks, it attains $61.1\%$ real-robot success versus $31.1\%$ for
single-candidate execution; under a fixed generation budget, the
corresponding rate for task-progress ranking is $35.6\%$.
\item We develop online forecast verification, which retains the selected
tactile forecast as an execution-time reference and triggers
corrective actions when persistent forecast--observation deviations are
detected. The full system attains an average real-robot success rate of
$72.2\%$ across the three tasks.
\end{itemize}

%% file: sections/02_related_work.tex
\section{Related Work}
\label{sec:related_work}

\subsection{Visual World Models and Action Generation}

Visual world models support planning by predicting action-conditioned future
observations. Existing approaches predict video in image space
\cite{finn2017visual}, learn probabilistic latent dynamics
\cite{chua2018deep,hansen2024tdmpc2}, or plan over pretrained visual features
\cite{zhou2025dinowm}; their planning utility depends on task progress being
observable in the predicted stream. A parallel line of work addresses action generation:
diffusion policies represent
multimodal action chunks \cite{chi2023diffusion}, while World action models
jointly generate video and continuous actions \cite{ye2026worldaction}. In the
latter, upcoming interactions are predicted primarily through visual signals,
leaving contact only indirectly represented. Variables that determine task
outcomes, such as contact timing, pressure distribution, and incipient slip,
may produce weak or delayed RGB signatures.
Appearance-based scores can therefore confuse a successful grasp
with a visually similar failure. HiTac-WAM augments each action proposal from a
pretrained world action model \cite{ye2026worldaction} with an explicit tactile
forecast.

\subsection{Visuo-Tactile Models and Tactile Prediction}

Tactile-conditioned policies fuse vision and touch to handle contact and
occlusion \cite{chen2023vtt}, relate touch to 3D
structure \cite{huang2025threedvitac}, and support bimanual or in-hand skills
\cite{gu2025tactilealoha,qi2023rotation}. Transferable representations,
robot-free interfaces, and human-centric pretraining broaden tactile supervision
\cite{yang2024unitouch,liu2025vitamin,zhang2026humancentric}, and tactile-augmented
VLA models incorporate touch into language-conditioned control
\cite{bi2026vlatouch,li2026atvla}. These methods use measured touch during
execution but do not explicitly predict the contact state of an unexecuted action.
Predictive tactile models optimize action sequences \cite{tian2019manipulation},
interpret future signals \cite{narang2020interpreting}, infer touch from visual
geometry \cite{ayad2024imagine2touch}, and estimate prospective slip
\cite{mandil2022action}. Measured tactile sequences also support estimates of
contact, force, slip, and grasp quality
\cite{declercq2022barometric,zhao2024tactilegrasp}. Together, these studies
establish the value of tactile state estimation and forecasting, but existing
predictive formulations typically either entangle contact state, interface
deformation, and slip risk within a single stream or model these quantities as
independent targets, without encoding their directed physical dependencies.
Recent models integrate tactile forecasting into a shared policy context.
ViTacFormer uses future tactile prediction to learn cross-modal representations
\cite{heng2025vitacformer}, Dream-Tac jointly predicts vision, touch, and action
\cite{lou2026dreamtac}, and UniTacVLA combines tactile latent prediction with
online correction \cite{zhang2026unitacvla}. Other recent systems use
visuo-tactile rollouts for planning
\cite{higuera2026visuotactileworldmodels},
couple tactile deformation with action prediction \cite{tian2026vtwam}, provide
high-frequency tactile correction
\cite{zheng2026omnivtavisuotactileworldmodeling}, or study rollout scaling
\cite{huang2026vitacworldscalingvisuotactileworld}. By contrast, HiTac-WAM
predicts a temporally resolved, hierarchical tactile world state for each
unexecuted action, explicitly evaluates the directed dependencies among
contact, deformation, and slip risk, and retains the selected forecast as an
execution-time reference.

\begin{figure*}[!t]
  \centering
  \includegraphics[width=0.98\textwidth,trim=2bp 220bp 0bp 258bp,clip]{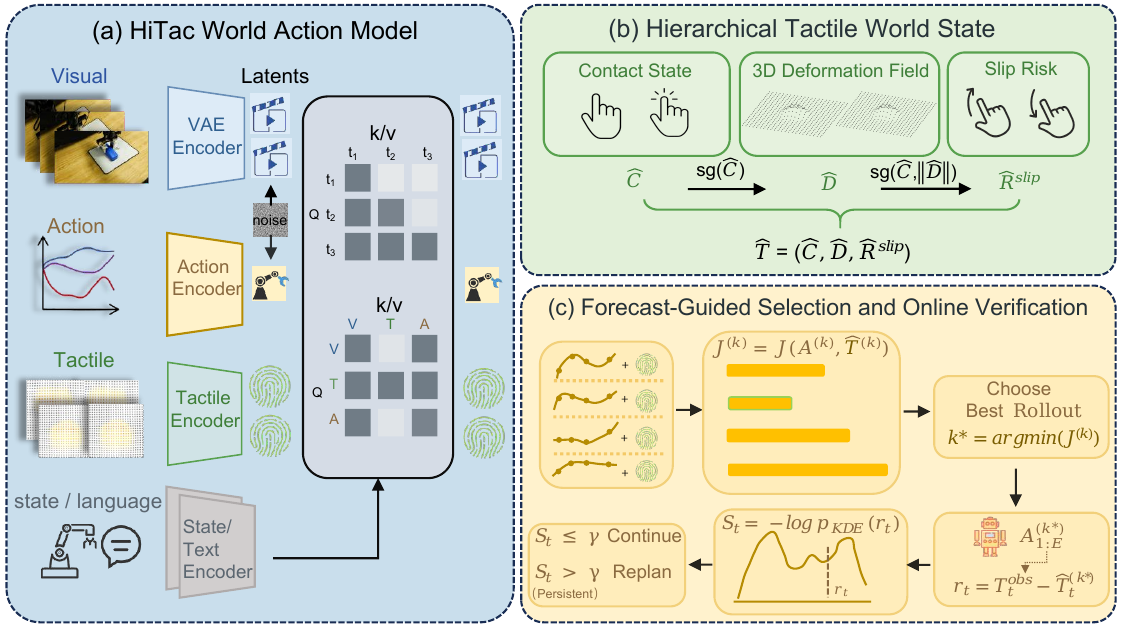}
  \caption{\textbf{Architecture of HiTac-WAM.} (a) A directed attention mask
  allows tactile queries to access the video--action context while isolating
  video and action queries from tactile history. (b) The hierarchy predicts
  bilateral contact, contact-conditioned 3D deformation, and contact-gated
  slip risk. (c) Candidate action--forecast pairs are ranked by $J^{(k)}$; the
  selected tactile forecast is retained as the execution-time reference, and
  persistent forecast deviations trigger corrective replanning.}
  \label{fig:architecture}
\end{figure*}

\subsection{Action Selection and Execution Monitoring}

Model-predictive and world-model planners rank candidates using learned
dynamics or scores in image, feature, or reward spaces
\cite{tian2019manipulation,zhou2025dinowm}, while diffusion policies model
multimodal action distributions \cite{chi2023diffusion}. KDPE instead uses
action-space density \cite{rosasco2025kdpe}. After selection, Reactive Diffusion
Policy and tactile primitives correct motion using measured touch
\cite{xue2025reactive,hogan2020tactile}. Multimodal anomaly detectors identify
execution anomalies \cite{ji2021multimodal}, while
task-specific recovery systems map failures to corrective actions
\cite{wu2018endowing}. HiTac-WAM links candidate selection and execution
monitoring by ranking candidates according to task progress and predicted
contact, deformation, and slip risk, and then retaining the selected forecast
as an execution-time reference; persistent deviations from this reference
trigger safe retreat and renewed candidate generation.

%% file: sections/03_method.tex
\section{Method}
\label{sec:method}

HiTac-WAM is a tactile world action model for contact-rich manipulation. Given
multimodal observations and a candidate action chunk, it forecasts the chunk's
hierarchical tactile consequences---contact state, contact-conditioned 3D
deformation, and slip risk---using a tactile branch that augments a pretrained
world action model \cite{ye2026worldaction}. At each replanning
step, forecasts for all candidate chunks guide ranking, and the selected
forecast is retained as an
execution-time reference. Fig.~\ref{fig:architecture} summarizes the
information flow.

\subsection{Directed Multimodal Context}

At control time $t$, all candidates share the observed history
$\mathcal H_t=\{\mathbf V_t,\mathbf q_t,\boldsymbol\tau_t,\ell\}$,
with $\mathbf V_t$ containing synchronized RGB views, $\mathbf q_t$ the robot
state, $\boldsymbol\tau_t$ the bilateral tactile history, and $\ell$ the
language instruction. Candidate $k\in\{1,\ldots,K\}$ is the $H$-step
action chunk
$\mathbf A_t^{(k)}=
\left[\mathbf a_t^{(k)},\ldots,\mathbf a_{t+H-1}^{(k)}\right]^\top
\in\mathbb R^{H\times7}$. The tactile branch $f_\theta$ predicts
\begin{equation}
\begin{aligned}
\widehat{\mathcal T}^{(k)}_{t+1:t+H}
&=f_\theta(\mathcal H_t,\mathbf A_t^{(k)}),\\
\widehat{\mathcal T}^{(k)}_{t+h}
&=\left(\widehat{\mathbf C}^{(k)}_{t+h},
\widehat{\mathbf D}^{(k)}_{t+h},
\widehat{\mathbf R}^{\mathrm{slip},(k)}_{t+h}\right).
\end{aligned}
\label{eq:model_io}
\end{equation}
At forecast offset $h$, action $\mathbf a_{t+h-1}^{(k)}$ is temporally aligned
with and conditions the tactile target at time $t+h$, and the same alignment
lets the selected forecast act as the execution-time reference. Together with
the generated video and action, $\widehat{\mathcal T}^{(k)}_{t+1:t+H}$ forms
the model's predicted world state: the rollout describes how the scene will
evolve, and the tactile hierarchy describes the contact consequences at the
interface.

The shared backbone receives visual, action, tactile, robot-state, and language
tokens. The pretrained FG-CLTP tactile encoder \cite{ma2026fgcltp} encodes the
bilateral tactile history into features that are projected into tactile tokens.
A directed attention mask
(Fig.~\ref{fig:architecture}(a)) constrains cross-modal information flow. Tactile
queries may attend to the tactile history and to the video--action context,
including its predicted future representations and aligned action tokens,
whereas visual and action queries cannot attend to tactile keys or values. With
rows indexing queries
and columns indexing keys in the order visual, action, and tactile, the
additive attention mask is
\begin{equation}
\mathbf M=
\begin{bmatrix}
\mathbf M_{VV} & \mathbf M_{VA} & -\infty\\
\mathbf M_{AV} & \mathbf M_{AA} & -\infty\\
\mathbf 0 & \mathbf 0 & \mathbf M_{TT}
\end{bmatrix},
\label{eq:mask}
\end{equation}
where the diagonal and visual--action blocks inherit the backbone's original
masks, the $-\infty$ blocks hide tactile keys from visual and action queries,
and the zero blocks give tactile queries full access to the visual and action
context.

During training, we update the tactile projections and queries, the
action--tactile fusion module, and the tactile prediction heads. Each tactile
forecast is therefore conditioned on the corresponding candidate action chunk
and its predicted visual future within the same model, rather than on an
external state estimate. Throughout, $h=1,\ldots,H$ and $s\in\{\mathrm
L,\mathrm R\}$ index forecast offsets and sensors, a hat marks a prediction,
and the superscript $\mathrm{obs}$ marks a causal online measurement; bold symbols
are vector- or grid-valued. We denote the feature
at offset $h$ for sensor $s$ by $\mathbf z_{t+h,s}^{(k)}$ and omit indices when
they are unambiguous.

\subsection{Explicit Hierarchical Tactile Forecast}
\label{sec:tactile_forecast}

The three outputs are physically ordered: contact,
interface deformation, and slip risk. Each deformation grid point contains two
in-plane displacement components and one surface-normal component, and
$\widehat C_{t+h,s}$ and $\widehat R^{\mathrm{slip}}_{t+h,s}$ denote the
scalar sensor-specific components of the bilateral outputs. Interface
deformation is meaningful only under contact, and slip risk depends on the
resulting interface state; HiTac-WAM therefore imposes a directed
contact-to-deformation-to-slip-risk hierarchy through three deterministic
prediction heads (Fig.~\ref{fig:architecture}(b)). Stop-gradient operations prevent downstream losses from
directly updating upstream heads, while the shared trainable tactile
representation couples the three predictions.

\begin{figure*}[!t]
  \centering
  \includegraphics[width=\textwidth,trim=18bp 318bp 12bp 258bp,clip]{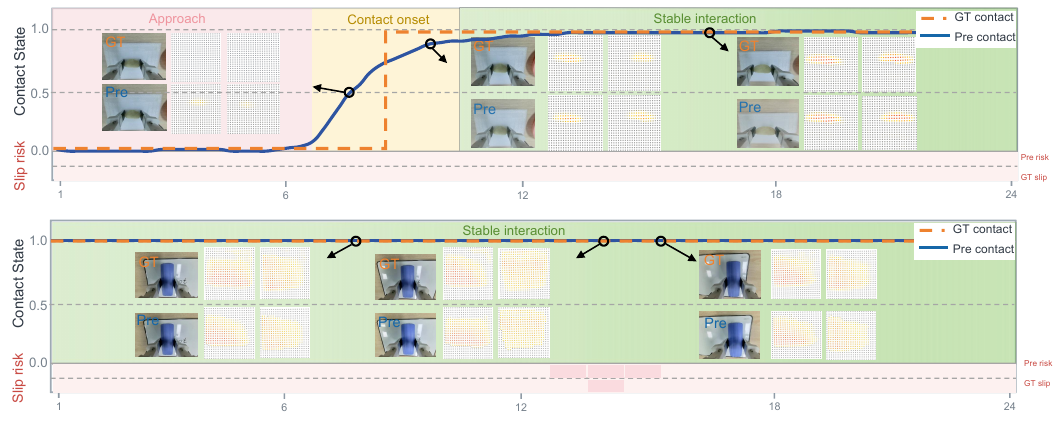}
  \caption{\textbf{Hierarchical Tactile World-State Prediction.}
  Representative 24-step forecasts for an approach-to-contact transition (top)
  and a stable-contact interval (bottom). Predicted contact-state estimates align
  with ground-truth transitions. Paired RGB and deformation snapshots show
  forecasts alongside the corresponding observations, while binary slip events
  are aligned with the predicted slip risk. Quantitative results for the three
  tasks are reported in
  Table~\ref{tab:tactile_prediction_main}.}
  \label{fig:tactile_prediction}
\end{figure*}

For a fixed candidate, forecast offset, and sensor, let $\mathbf z$ and
$\mathbf a$ denote the corresponding tactile feature and aligned action. Let
$\mathbf z^C$ denote an intermediate transition feature produced within the
contact head; for compactness, $f_C$ below denotes its scalar logit readout.
The current deformation is
$\mathbf D_t=\mathbf P_t-\mathbf P_{\mathrm{base}}$, with $\mathbf P_t$ and
$\mathbf P_{\mathrm{base}}$ denoting the observed and undeformed sensor grids.
The training target at offset $h$ is
$\Delta\mathbf D_{t+h,s}=\mathbf D_{t+h,s}-\mathbf D_{t,s}
=\mathbf P_{t+h,s}-\mathbf P_{t,s}$. The contact head affects downstream
prediction in two ways: its detached transition feature
$\operatorname{sg}(\mathbf z^C)$ conditions the deformation head, whereas the
detached contact-state estimate $\operatorname{sg}(\widehat C)$ conditions the
slip head. In the conditioning
interventions of Section~\ref{sec:results}, $\operatorname{sg}(\mathbf z^C)$
is replaced by a learned contact embedding indexed by the intervened value,
so that ground-truth, predicted, shuffled, and zero contact differ only in
the conditioning signal. The predicted contact-state estimate $\widehat C$
gates both physical outputs, with $\odot$ denoting its broadcast elementwise
application to the sensor grid. To simplify notation, we omit $(k,t+h,s)$ below and evaluate the
three heads in physical order:
\begin{subequations}
\label{eq:heads}
\begin{align}
\widehat C
&=\sigma\!\left(f_C(\mathbf z,\mathbf a)\right),
\label{eq:contact_head}
\\
\Delta\widehat{\mathbf D}
&=f_D(\mathbf z,\operatorname{sg}(\mathbf z^C),\mathbf a),\notag\\
\widehat{\mathbf D}
&=\widehat C\odot
\left(\mathbf D_t+\Delta\widehat{\mathbf D}\right),
\label{eq:deformation_head}
\\
\widehat p^{\mathrm{slip}}
&=\sigma\!\Bigl(f_R\!\bigl(
\mathbf z,
\operatorname{sg}(\widehat C),
\operatorname{sg}\!\left(
\left\|\mathbf D_t+\Delta\widehat{\mathbf D}\right\|_2
\right)
\bigr)\Bigr),\notag\\
\widehat R^{\mathrm{slip}}
&=\widehat C\,\widehat p^{\mathrm{slip}}.
\label{eq:slip_head}
\end{align}
\end{subequations}
Here, $\sigma$ is the sigmoid function, and $\operatorname{sg}$ denotes
stop-gradient. The ungated future deformation
$\mathbf D_t+\Delta\widehat{\mathbf D}$ conditions slip prediction through
its magnitude; Fig.~\ref{fig:architecture}(b) depicts this dependency
schematically. The deformation and slip losses supervise
$\Delta\widehat{\mathbf D}$ and $\widehat p^{\mathrm{slip}}$, respectively,
whereas downstream selection and control use the contact-gated outputs
$\widehat{\mathbf D}$ and $\widehat R^{\mathrm{slip}}$.
Fig.~\ref{fig:tactile_prediction} presents representative forecasts.

Let $\mathbf C$, $\Delta\mathbf D$, and $\mathbf R^{\mathrm{slip}}$ denote the
contact, deformation-increment, and slip-event targets stacked over the
forecast horizon and both sensors. We train the three outputs jointly using
the objective
\begin{equation}
\begin{aligned}
\mathcal L&=\mathcal L_C+\mathcal L_D+\beta\mathcal L_R,\\
\mathcal L_C&=\operatorname{BCE}(\widehat{\mathbf C},\mathbf C)
+\lambda_{\mathrm{on}}\mathcal L_{\mathrm{onset}}
+\lambda_{\mathrm{rel}}\mathcal L_{\mathrm{release}},\\
\mathcal L_D&=\frac{\sum_{h,s}C_{t+h,s}\,
\rho_\delta\!\left(\Delta\widehat{\mathbf D}_{t+h,s}
-\Delta\mathbf D_{t+h,s}\right)}
{\sum_{h,s}C_{t+h,s}+\epsilon_D},\\
\mathcal L_R&=\operatorname{WBCE}
(\widehat{\mathbf p}^{\mathrm{slip}},\mathbf R^{\mathrm{slip}}).
\end{aligned}
\label{eq:total_loss}
\end{equation}
Here, $\rho_\delta$ denotes the elementwise Huber penalty, averaged over
the tactile grid and coordinate dimensions. The contact targets mask the
deformation loss to forecast offsets and sensors with ground-truth contact,
$\epsilon_D>0$ prevents division by zero in contact-free windows, and
$\operatorname{WBCE}$ denotes class-weighted binary cross-entropy. The onset
and release terms constrain the timing of contact transitions. Because the
inputs from the upstream heads to the slip head are detached, $\mathcal L_R$
does not directly update the contact or deformation heads.
Section~\ref{sec:experiments} provides the loss weights and optimization
settings.

\subsection{Forecast-Guided Candidate Selection}
\label{sec:selection}

At each replanning step, the pretrained world action model samples $K$
stochastic candidates from the shared visual, robot-state, and language
context using independent noise draws. Kinematic checks
then discard invalid action chunks, yielding
$\mathcal K_{\mathrm{valid}}\subseteq\{1,\ldots,K\}$.

A single batched evaluation of $f_\theta$ predicts a tactile trajectory for
every $k\in\mathcal K_{\mathrm{valid}}$. Each forecast is reduced to normalized
rollout-level costs $J_C^{(k)}$, $J_D^{(k)}$, and $J_R^{(k)}$, which
respectively measure phase-inconsistent contact, excessive or bilaterally
imbalanced deformation, and slip risk; $J_R^{(k)}$ aggregates peak and average
risk over the bilateral forecast. The task score
$\rho_{\mathrm{task}}^{(k)}$ measures predicted progress from the candidate's
predicted visual rollout. The combined selection score therefore penalizes
harmful contact rather than contact itself. Candidate selection
(Fig.~\ref{fig:architecture}(c)) follows
\begin{equation}
\begin{aligned}
J^{(k)}&=-w_{\mathrm{prog}}\rho_{\mathrm{task}}^{(k)}
+w_CJ_C^{(k)}+w_DJ_D^{(k)}+w_RJ_R^{(k)},\\
k^*&=\arg\min_{k\in\mathcal K_{\mathrm{valid}}}J^{(k)}.
\end{aligned}
\label{eq:risk_score}
\end{equation}
Before weighting, all four score components are normalized to $[0,1]$.
After selecting $k^*$, the controller begins executing
$\mathbf A_t^{(k^*)}$ and retains
$\widehat{\mathcal T}_{t+1:t+H}^{(k^*)}$ as the temporally aligned reference
described next.

\subsection{Online Forecast Verification}
\label{sec:online_verification}

The controller executes a length-$E$ prefix of the selected action chunk,
where $E\leq H$, before replanning. At each executed offset
$i=1,\ldots,E$, the observation $\mathcal T^{\mathrm{obs}}_{t+i}$, obtained
after executing $\mathbf a_{t+i-1}^{(k^*)}$, is compared with the aligned
forecast $\widehat{\mathcal T}^{(k^*)}_{t+i}$:
\begin{equation}
\begin{aligned}
\mathbf r_{t+i}
&=d\!\left(\mathcal T^{\mathrm{obs}}_{t+i},
\widehat{\mathcal T}^{(k^*)}_{t+i}\right),\\
S_{t+i}
&=-\log p_{\mathrm{KDE}}^{(\mathrm{task})}
\!\left(\operatorname{std}_{\mathrm{task}}(\mathbf r_{t+i})\right),
\end{aligned}
\label{eq:monitor}
\end{equation}
The discrepancy function $d$ concatenates the bilateral absolute errors for
contact and slip with the deformation Top-20 Z MAE.
$\operatorname{std}_{\mathrm{task}}$ standardizes each component using
task-specific validation statistics. The online contact and slip measurements
are computed causally from the same current and historical signals used to
generate the recorded labels, without using future frames.

For each task, a separate kernel density estimator (KDE) is fitted to the
standardized residuals from successful validation episodes. A forecast
deviation is declared only when $S_{t+i}$ remains above the calibrated
threshold $\gamma_{\mathrm{task}}$ for the configured number of consecutive
executed offsets. The persistent deviation triggers a recovery
mechanism (Fig.~\ref{fig:architecture}(c)): the controller aborts the unexecuted remainder of the prefix,
returns the robot to a safe state, acquires a new observation, and restarts
candidate generation.

%% file: sections/04_experiments.tex
\section{Experiments}
\label{sec:experiments}

The experiments answer four questions.
Can HiTac-WAM predict contact, 3D deformation, and slip risk across the action
horizon (\textbf{Q1})? Does the directed hierarchy improve these predictions,
and how much do the video and action outputs change (\textbf{Q2})? Do
candidate-specific tactile forecasts improve action selection over stochastic
sampling, task-progress ranking, and current-touch control (\textbf{Q3})? How
does the selected tactile forecast support online forecast verification during
execution (\textbf{Q4})?

\subsection{Platform, Data, and Training}

\begin{figure}[!b]
  \centering
  \includegraphics[width=0.98\linewidth,trim=210bp 210bp 0bp 290bp,clip]{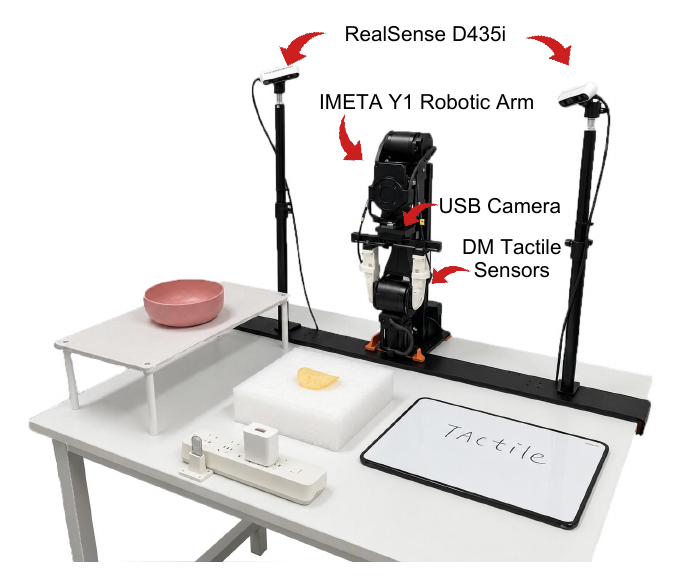}
  \caption{\textbf{Real-Robot Experimental Platform.} The platform comprises an
  IMETA-Y1 robot, two Intel RealSense D435i cameras, one USB camera, and two
  DM-Tac W2 tactile sensors mounted on the gripper. The workspace contains the
  objects used in chip grasping, blackboard erasing, and USB insertion.}
  \label{fig:experimental_platform}
\end{figure}

All experiments use the IMETA-Y1 robot shown in
Fig.~\ref{fig:experimental_platform}, equipped with bilateral DM-Tac W2 tactile
sensors and three synchronized RGB views. The hardware is fixed across methods
and data splits.

We evaluate chip grasping, blackboard erasing, and USB insertion. For each task,
200 complete episodes are partitioned before temporal-window
extraction into 160 training, 20 validation, and 20 test episodes; no window
crosses a data split. Each stream includes bilateral contact and slip labels.
Each 3D deformation annotation represents the marker-wise displacement from the
no-contact state of the corresponding sensor.

The streams are synchronized at 30 Hz. Each sample comprises tactile history at
relative offsets $[-12,-6,-2,0]$, actions at offsets $[0,\ldots,23]$, and
tactile targets at offsets $[1,\ldots,24]$; tactile observations form two
sensor grids. The tactile prediction modules are trained
separately for each task for 10,000 steps on 8 NVIDIA H100 GPUs. Contact-onset and
contact-release losses have weights $2$ and $1$, respectively, and positive
slip events have weight $30$. The slip-loss coefficient is $0$ for the first
3,000 steps and $0.01$ thereafter. Optimization uses AdamW with a learning
rate of $10^{-4}$ and a global batch size of 64.

\subsection{Prediction Metrics and Baselines}

Contact metrics comprise F1 and onset and release MAEs in 30-Hz frames.
For future frames with ground-truth contact, we report denormalized 3D
displacement L2, Top-20 Z MAE, and the predicted-to-ground-truth deformation
norm ratio. Top-20 Z MAE averages the absolute surface-normal displacement error
over the 20 points with the largest ground-truth absolute Z deformation; the
onset variant uses contact-onset frames. Because positive slip events occur in
only $1.39\%$ of frames, we use AUPRC, the area under the precision--recall
curve. Learned perceptual image patch similarity (LPIPS) and normalized
action-trajectory MAE measure video and action preservation, respectively.

Validation data are used exclusively for
checkpoint selection, KDE fitting, and threshold calibration.

\subsection{Closed-Loop Protocol}

DreamZero executes one sampled action chunk and corresponds to the same world
action model without the tactile pathway. Reactive Tactile uses current
touch without tactile foresight, whereas the task-progress baseline ranks four
stochastic candidates without tactile forecasts. Selection ranks four
stochastic candidates using forecasts of contact, deformation, and slip risk.
Full HiTac-WAM also retains the selected forecast as an execution-time
reference, with persistent deviations triggering corrective actions. Selection
and Full HiTac-WAM combine worst-case and mean slip risk and use
$(w_{\mathrm{prog}},w_C,w_D,w_R)=(1.0,1.0,1.0,0.5)$
in~\eqref{eq:risk_score}.
Table~\ref{tab:closed_loop_main} reports the 360-trial main comparison;
Section~\ref{sec:results_selection} provides a fixed-budget $K=4$ ranking
comparison: the tactile-forecast arm reuses the Selection trials from
Table~\ref{tab:closed_loop_main}, and the task-progress arm adds a 90-trial set
with the same task distribution, generation budget, and validity rules;
candidate seeds are matched between the two arms, but their executions remain
independent.

Each method is evaluated in 30 trials per task. Chip grasping is successful if
an intact chip reaches the target plate without being dropped and remains
stable for at least 3 s. Blackboard erasing requires effective contact along
the prescribed path and removal of more than $80\%$ of the target region. USB
insertion requires full insertion without sustained jamming and activation of
the connector LED.

KDE settings and task-specific thresholds are calibrated using successful
validation episodes and fixed before evaluation. The executed prefix comprises
8 steps for all three tasks. Three consecutive
scores above the threshold trigger correction, with at most two corrective
attempts per trial. Detector statistics are estimated from post-hoc
inspection of trial recordings rather than exhaustive anomaly annotation.

Held-out sequences measure forecast quality, whereas real-robot trials measure
how the forecasts affect executed actions. The fixed-budget $K=4$ comparison is
descriptive.

%% file: sections/05_results.tex
\section{Results}
\label{sec:results}

\input{tables/closed_loop_main_results}

\subsection{Tactile World-State Prediction}

\input{tables/tactile_prediction_main}

Table~\ref{tab:tactile_prediction_main} answers Q1 across the three tasks.
Contact F1 spans $0.908$--$0.931$ (mean $0.921$),
with mean onset and release errors of $2.1$ and $2.3$ frames at 30 Hz;
individual timing errors span $1.8$--$2.6$ frames, or roughly $60$--$87$ ms.
The 3D
displacement L2 and Top-20 Z MAE are $0.054$--$0.062$ mm and
$0.071$--$0.081$ mm, respectively, while norm ratios of $0.93$--$0.99$
indicate comparable predicted and measured deformation magnitudes. Slip AUPRC
is $0.232$--$0.258$ (mean $0.247$), well above the $0.0139$ chance level;
because slip events occupy only $1.39\%$ of frames, we treat this output as a
ranking signal rather than a calibrated alarm. The task-wise values remain
similar across transient and sustained contact regimes, so no single task
drives the means.

Fig.~\ref{fig:tactile_prediction} shows predicted contact-state estimates
tracking the ground-truth transitions, deformation snapshots matching the paired RGB
observations, and slip risk remaining near zero during free motion and rising
around labeled slip events; the markers are binary because the labels denote
events rather than continuous probabilities.

\subsection{Hierarchy and Conditioning}
\label{sec:results_hierarchy}

\input{tables/hierarchy_ablation}

\begin{figure}[!b]
  \centering
  \includegraphics[width=0.94\columnwidth]{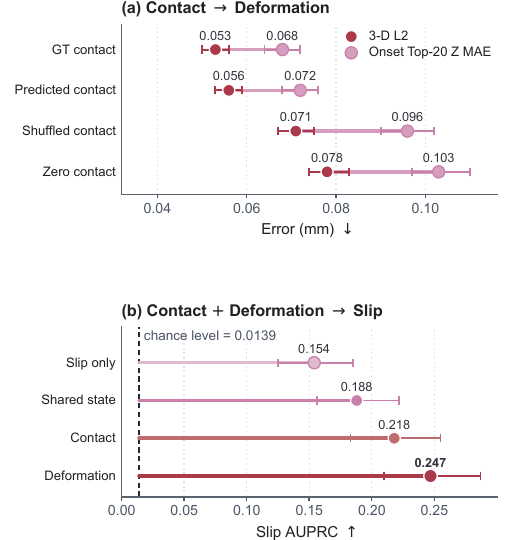}
  \caption{\textbf{Hierarchical Conditioning Diagnostics.}
  (a) Contact-intervention 3D L2 and onset Top-20 Z MAE, with episode-paired
  bootstrap 95\% CIs. (b) Slip AUPRC under cumulative conditioning; segments
  show lift above prevalence ($0.0139$), with stratified episode-bootstrap
  95\% CIs.}
  \label{fig:conditioning_diagnostics}
\end{figure}

Table~\ref{tab:hierarchy_ablation} answers Q2: the directed hierarchy performs
best across all four metrics---$0.921$ F1, $0.056$ mm 3D L2 ($17.6\%$ below
the deformation-only predictor), $0.072$ mm Top-20 Z, and a $0.97$ norm ratio.
The intermediate rows explain why: independent heads do not consistently
surpass the single-output predictors (F1: $0.896$ vs. $0.903$; 3D L2:
$0.064$ vs. $0.068$ mm), so sharing alone is insufficient. Contact
conditioning ($0.915$ F1 and $0.059$ mm) recovers most of the improvement,
and the full hierarchy provides the remaining gain.

With the checkpoint fixed, Fig.~\ref{fig:conditioning_diagnostics}(a)
probes contact conditioning by corrupting the contact input to the deformation
head. The 3D L2 rises monotonically across ground-truth, predicted, shuffled,
and zero contact ($0.053$, $0.056$, $0.071$, and $0.078$ mm); the endpoint
intervals do not overlap. Conditioning on predicted contact nearly matches
ground-truth conditioning at inference, while the graded degradation confirms
that the deformation head uses this signal. Panel (b) shows the same staircase
for slip: AUPRC rises from $0.154$ (slip only) to $0.188$ (shared hidden state),
$0.218$ (contact conditioning), and $0.247$ (contact and deformation), a
$60.4\%$ relative gain. Current-grid and base-grid temporal priors give 3D
L2/Top-20 Z errors of $0.079/0.104$ and $0.073/0.099$ mm, both worse than the
hierarchy's $0.056/0.072$ mm. Relative to the world action model without tactile prediction,
video LPIPS and action MAE are higher by $1.4\%$ and $0.9\%$, respectively.

\subsection{Forecast-Guided Action Selection}
\label{sec:results_selection}

\begin{figure}[!t]
  \centering
  \includegraphics[width=3.0in]{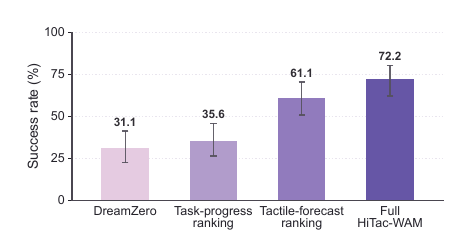}
  \caption{\textbf{From Sampling to Ranking to Verification.} Observed success
  rates. At the fixed $K{=}4$ budget, tactile-forecast ranking reuses the
  Selection trials from Table~\ref{tab:closed_loop_main}, whereas task-progress
  ranking uses an additional independent 90-trial set. DreamZero executes one
  candidate; error bars show two-sided 95\% Wilson intervals.}
  \label{fig:ranking_comparison}
\end{figure}

\begin{figure}[!t]
  \centering
  \includegraphics[width=0.88\columnwidth]{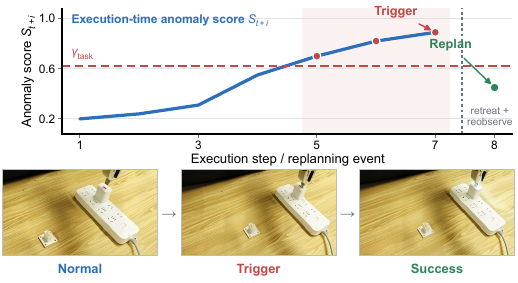}
  \caption{\textbf{Online Forecast Verification during Execution.} Anomaly-score
  trace recorded in a real execution trial ($E{=}8$ prefix); three consecutive
  scores above $\gamma_{\mathrm{task}}$ trigger safe retreat and replanning.}
  \label{fig:online_forecast_verification}
\end{figure}

Q3 asks whether the forecast changes which action is executed. The gains over
DreamZero are largest on chip grasping and USB insertion, where contact
determines the outcome. Under
the fixed generation budget in
Fig.~\ref{fig:ranking_comparison}, tactile-forecast ranking reuses the Selection
trials from Table~\ref{tab:closed_loop_main}, attaining $61.1\%$, whereas
task-progress ranking attains $35.6\%$ on an additional matched-seed 90-trial
set. Their two-sided 95\% Wilson intervals do not overlap ($[50.8,70.5]$ and
$[26.4,45.8]$).

\subsection{Online Forecast Verification}

Q4 examines execution-time use of the retained forecast;
Fig.~\ref{fig:online_forecast_verification} shows a real replanning trigger.
In Table~\ref{tab:closed_loop_main}, Full HiTac-WAM attains $72.2\%$ success,
compared with $61.1\%$ for Selection.
Post-hoc inspection of the 90
Full-configuration trials identified 18 cases in which verification triggered
replanning. Sixteen of these cases coincided with execution anomalies, 12 of
which were followed by successful task completion; the remaining two were
false triggers with no observed effect on the outcome. At most three of the 21
untriggered failures involved an undetected anomaly.

\FloatBarrier

%% file: tables/closed_loop_main_results.tex
\begin{table}[!t]
\caption{Success rates (\%) in the main real-robot comparison (30 trials per
method--task pair; 360 total).}
\label{tab:closed_loop_main}
\centering
\footnotesize
\setlength{\tabcolsep}{2.5pt}
\renewcommand{\arraystretch}{1.12}
\begin{tabularx}{\columnwidth}{l Y Y Y Y}
\toprule
Method & Chip Grasping & Blackboard Erasing & USB Insertion & Average \\
\midrule
DreamZero & 33.3 & 50.0 & 10.0 & 31.1 \\
Reactive Tactile & 40.0 & 60.0 & 20.0 & 40.0 \\
HiTac-WAM Selection & 70.0 & 73.3 & 40.0 & 61.1 \\
\rowcolor{HiTacDeploy}
\textbf{Full HiTac-WAM} & \textbf{76.7} & \textbf{90.0} & \textbf{50.0}
& \textbf{72.2} \\
\bottomrule
\end{tabularx}
\end{table}

%% file: tables/tactile_prediction_main.tex
\begin{table*}[!t]
\caption{Future tactile prediction. Timing errors are in 30-Hz frames and 3D
errors in millimeters; Norm is the predicted-to-ground-truth deformation norm
ratio. Means are unweighted across tasks.}
\label{tab:tactile_prediction_main}
\centering
\footnotesize
\setlength{\tabcolsep}{4.5pt}
\renewcommand{\arraystretch}{1.12}
\begin{tabularx}{\textwidth}{L d{1.3} d{1.3} d{1.3} d{1.3} d{1.3} d{1.3} d{1.3}}
\toprule
& \multicolumn{3}{c}{Contact}
& \multicolumn{3}{c}{3D deformation}
& \multicolumn{1}{c}{Slip risk} \\
\cmidrule(lr){2-4}\cmidrule(lr){5-7}\cmidrule(l){8-8}
Task & \multicolumn{1}{c}{Contact F1 $\uparrow$}
& \multicolumn{1}{c}{Onset MAE $\downarrow$}
& \multicolumn{1}{c}{Release MAE $\downarrow$}
& \multicolumn{1}{c}{3D L2 (mm) $\downarrow$}
& \multicolumn{1}{c}{Top-20 Z MAE (mm) $\downarrow$}
& \multicolumn{1}{c}{Norm ratio $\to 1$}
& \multicolumn{1}{c}{AUPRC $\uparrow$} \\
\midrule
Chip grasping & 0.924 & 2.0 & 2.3 & 0.058 & 0.075 & 0.960 & 0.232 \\
Blackboard erasing & 0.908 & 2.4 & 2.6 & 0.062 & 0.081 & 0.930 & 0.251 \\
USB insertion & 0.931 & 1.8 & 2.1 & 0.054 & 0.071 & 0.990 & 0.258 \\
\midrule
\rowcolor{HiTacDeploy}
\textbf{Mean} & 0.921 & 2.1 & 2.3 & 0.058 & 0.076 & 0.960 & 0.247 \\
\bottomrule
\end{tabularx}
\end{table*}

%% file: tables/hierarchy_ablation.tex
\begin{table}[!b]
\caption{Hierarchy ablation under matched training budgets. Errors are in
millimeters, and Norm is the predicted-to-ground-truth deformation norm ratio.
The deformation-only predictor is supervised on contact frames; independently
trained rows need not match Table~\ref{tab:tactile_prediction_main}. Boldface
and underlining mark the best and second-best values.}
\label{tab:hierarchy_ablation}
\centering
\footnotesize
\setlength{\tabcolsep}{2.5pt}
\renewcommand{\arraystretch}{1.12}
\begin{tabularx}{\columnwidth}{l Y Y Y Y}
\toprule
Model & F1 $\uparrow$ & 3D L2 $\downarrow$ & Top-20 Z $\downarrow$ & Norm $\to 1$ \\
\midrule
Contact-only & 0.903 & --- & --- & --- \\
Deformation-only & --- & 0.068 & 0.090 & 0.82 \\
\addlinespace[1pt]
Independent heads & 0.896 & 0.064 & 0.083 & 0.87 \\
Contact-conditioned
& \underline{0.915} & \underline{0.059} & \underline{0.076}
& \underline{0.94} \\
\rowcolor{HiTacDeploy}
\textbf{Directed hierarchy}
& \textbf{0.921} & \textbf{0.056} & \textbf{0.072} & \textbf{0.97} \\
\bottomrule
\end{tabularx}
\end{table}

%% file: sections/07_conclusion.tex
\section{Conclusion}
\label{sec:conclusion}

HiTac-WAM models the tactile consequences of each candidate action chunk as a
directed contact-to-deformation-to-slip-risk hierarchy that reflects their
physical dependencies. For tactile prediction, the model attains a mean
contact F1 of $0.921$, onset and release errors of $2.1$ and $2.3$ frames, 3D
displacement L2 of $0.058$ mm, and slip AUPRC of $0.247$ across the three
tasks. Matched-budget ablations show that shared multi-output prediction alone
is insufficient: the directed hierarchy reduces 3D displacement L2 error by
$17.6\%$ relative to the deformation-only predictor and improves slip AUPRC by
$60.4\%$ relative to the slip-only predictor. Relative to the world action
model without tactile prediction, video LPIPS and action MAE are higher by
$1.4\%$ and $0.9\%$, respectively. For action selection, ranking guided by
these hierarchical forecasts attains $61.1\%$ success versus $31.1\%$ for
single-candidate execution; under a fixed generation budget, it attains the
same $61.1\%$ versus $35.6\%$ for task-progress ranking, with non-overlapping
confidence intervals. With online forecast verification, the full system
attains $72.2\%$. Together, the hierarchy ablations show that organizing
tactile predictions according to their physical dependencies improves
predictive accuracy. The real-robot results demonstrate the utility of keeping
each forecast attached to its candidate action: the forecast is consulted
before execution to rank candidates and retained during execution as the
reference against which reality is checked.

Despite these results, forecast quality on model-generated action chunks is
not evaluated separately, and predicted slip risk is not calibrated for online
detection. Future work will directly evaluate such candidates and controlled
anomalies while exploring on-policy adaptation, uncertainty-calibrated slip
forecasting, and learned corrective policies for long-horizon contact-rich
robotic manipulation tasks.